\documentclass[runningheads]{llncs}
\usepackage{graphicx}
\usepackage{amsmath}

\begin{document}
\title{Deep Transparent Prediction through Latent Representation Analysis}
\titlerunning{Deep Transparent Prediction}
%
\author{D. Kollias \inst{1} \and N. Bouas \inst{1} \and
Y. Vlaxos \inst{1} \and V. Brillakis \inst {1} \and M. Seferis  \inst{1}  \and I. Kollia \inst{1} \and\\ L. Sukissian \inst {2} \and  J. Wingate \inst {3} \and S. Kollias \inst{1,2,3}} 
\authorrunning{Kollias et al.}
%
\institute{School of Electrical and Computer Engineering, National Technical University of Athens, Athens, Greece \and 
GRNET National Infrastructures for Research and Technology, Athens, Greece \and 
School of Computer Science, University of Lincoln, Lincoln, UK} 
%

%
\maketitle              
\begin{abstract}
The paper presents a novel deep learning approach, which extracts latent information from trained Deep Neural Networks (DNNs) and derives concise representations that are analyzed in an effective, unified way for prediction purposes. It is well known that DNNs are capable of analyzing complex data; however, they lack transparency in their decision making, in the sense that it is not straightforward to justify their prediction, or to visualize the features on which the decision was based. Moreover, they generally require large amounts of data in order to learn and become able to adapt to different environments. This makes their use difficult in healthcare, where trust and personalization are key issues. 
Transparency combined with high prediction accuracy are the targeted goals of the proposed approach. It includes both supervised DNN training and unsupervised learning of latent variables extracted from the trained DNNs. Domain Adaptation from multiple sources is also presented as an extension, where the extracted latent variable representations are used to generate predictions in other, non-annotated, environments. Successful application is illustrated through a large experimental study in various fields:  prediction of Parkinson's disease from MRI and DaTScans; prediction of COVID-19 and pneumonia from CT scans and X-rays; optical character verification in retail food packaging.

\end{abstract}
\section{Introduction}
Over the last few years, Deep Learning (DL) and Deep Neural Networks (DNNs)  have been successfully applied to numerous applications and domains due to the availability of large amounts of labeled data, including healthcare prediction, visual analysis and recognition \cite{ref1,ref2,ref3}. 

Transfer learning (TL) \cite{ref4} has been the main approach to train Deep Neural Networks when only small amounts of annotated data are available. TL uses networks previously trained with large datasets (even of generic patterns) and fine-tunes the whole, or parts of them, using the small training datasets. A serious problem is related to TL: as the DNN learns to make predictions in the new dataset, it tends to forget the old data that are not used in the retraining procedure; this is known as `catastrophic forgetting'. 
Moreover, when deploying a pre-trained model to a real-life application, the assumption is that both the source (training set) and the target (application-specific) one are drawn from the same distribution. When this assumption is violated, the DL model trained on the source domain will not generalize well on the target domain due to the distribution differences between the two domains; this is known as domain shift. Learning a discriminative model in the presence of domain shift between  target datasets is known as Domain Adaptation (DA) \cite{ref5} and is targeted when dealing with non-annotated data.

Recent research has focused on extracting trained DNN representations and using them for classification purposes \cite{ref6}, either by an auto-encoder methodology, or by monitoring neuron outputs in the convolutional or/and fully connected network layers \cite{ref7,ref8}. Such developments are exploited in this paper, proposing a novel approach that is able to generate unified prediction models, providing transparency and visualization of their decision making process in a variety of application domains.  
At first, we extract appropriate internal features, say features $\textbf v$, from a DNN model trained with some dataset of interest. Using a clustering methodology, we generate concise representations, say $\textbf c$, of these features. Using these representations and the nearest neighbour criterion, we can then predict, in an efficient and transparent way, the class of new data.   

Combining DNN training and clustering has been a topic of recent research. Surveys, focusing on different clustering methodologies and different combination ways can be found in \cite{ref27,ref29}; specific combinations can also be found in \cite{ref28,ref30,ref31}. Here, our aim is to derive the unified latent representation and prediction framework, illustrating its successful use, especially in medical applications. The framework of interweaving DNNs and clustering has also been examined in our former publication \cite{ref32}.

Next, we present a new methodology that alleviates the `catastrophic forgetting' problem by generating a unified model over different datasets. According to this methodology, we apply the  originally trained DNN to a new dataset deriving a corresponding set of representations, through which we train a new DNN. From the latter DNN, we extract a new set of features, say $\textbf v'$ and a concise representation $\textbf c'$. A unified prediction model is then produced by merging the $\textbf c$ and $\textbf c'$ representation sets. Having achieved high precision and recall metrics in the derivation of each one of these representations ensures that the generated unified model provides high prediction accuracy in the derived representation space. This alternative prediction is of great significance in the case of new non-annotated data, since it provides a transparent way for prediction; it is shown that it can also create richer representations of the prediction problem. We then use the extracted latent variable representations from trained DNNs in multiple sources, so as to generate predictions in other environments. 

The proposed methodology is applied to a variety of applications, focusing on medical imaging for healthcare, but also on other applications where image analysis is used for anomaly prediction. In the latter case we focus on quality control in retail food packaging, based on real images provided by large supermarkets \cite{ref9}. In the former case we focus: a) on prediction of Parkinson’s, based on datasets of MRI and DaTScans, either created in collaboration with the Georgios Gennimatas Hospital (GGH) in Athens \cite{ref10}, or provided by the PPMI study sponsored by M. J. Fox for Parkinson’s Research \cite{ref11}, b) on prediction of COVID-19, based on CT chest scans and x-rays, either public, or aggregated by GRNET in collaboration with the Hellenic Ministry of Health.  
  
The novel contributions of the paper are the following:
i)	we develop a novel unsupervised learning approach, extracting latent variables from trained DNNs, which, after appropriate clustering, provide unified concise representations that can be analyzed in an efficient and transparent way for prediction;
ii)	each concise representation set is linked to the respective input data (i.e., medical images, or scans, or other information); we are, therefore, able to show - to the (medical) experts and users/patients - which were the main (similar) cases on which the provided prediction/diagnosis was based. It is then up to the experts/users to decide whether they trust (this basis of) the diagnosis, or not;
iii) we present a DA framework, from multiple sources, which uses DNNs and extracted representations from annotated datasets, so as to generate predictions in other, non-annotated data, collected in different environments; 
ii)	we apply the proposed approach to the following application domains:  i) for unified Parkinson's disease prediction, over different datasets, based on medical imaging, ii) for effective prediction of COVID-19 from CT chest scan series, or from x-rays, iii) for optical character verification on food retail packaging, based on DA among different datasets.

\vspace{-0.2cm}
\section {Related work}

Related research in DNN representations' generation  includes Generative Adversarial Network (GAN) formulations \cite{ref12} - to provide data augmentation and improve generalization of the process - and capsules \cite{ref13,ref14} – to include feedback and model structuring in the generated models.  Moreover, the derived DL framework can be expressed in a multi-objective optimization framework, so that the approach is coupled with a robust analysis methodology \cite{ref15}.

In recent years, many single source domain adaptation methods have been proposed. Discrepancy-, Adversarial- and Reconstruction- based approaches are the three primary domain adaptation approaches currently being applied to address the distribution shift \cite{ref5}. Discrepancy-based approaches rely on aligning the distributions in order to minimize the divergence between them. The most commonly used discrepancy based methods are Maximum Mean Discrepancy (MMD) \cite{ref16} and Correlation Alignment (CORAL) \cite{ref17}. Adversarial-based approaches minimize the distance between the source and the target distributions using domain confusion, an adversarial method used in GANs. Another class of approaches known as Reconstruction-based approaches create a shared representation between the source and target domains whilst preserving the individual characteristics of each domain. Rather than minimizing the divergence, \cite{ref18} learns joint representations that classify the labeled source data and at the same time reconstruct the target domain. Moreover, multi-source domain adaptation techniques need to handle both domain alignment between source and target domains, along with alignment between multiple available sources. 

In this paper we briefly present a DA approach that we will use in an extended pipeline, following the derivation of the unified representation framework. In this, we minimize the feature discrepancy through implementing a new loss function that includes both MMD and CORAL losses for improved generalization. Additionally, a Class Activation Mapping (CAM) component \cite{ref19} adds an extra step to the algorithm that provides a visualization of which areas in the data contributed the most to the decision-making process. CAM adds insight into the process of CNN interpretability and explainability by overlaying a heat map over the data to demonstrate where the model is paying more attention in its decision-making process. 

\vspace{-0.2cm}
\section{The Proposed Methodology}\label{sec3}

\subsection{The Extracted Features from Deep Neural Networks}

Our approach starts by training a CNN or CNN-RNN, to predict the status of data samples. Let us assume that we perform analysis of medical images, e.g., MRI and scans, collected in a specific medical centre, or hospital. 
 As in \cite{ref2} we consider a CNN part that has a well-known structure, such as ResNet-50, generally composed of convolutional and pooling layers, followed by one, or two fully-connected layers. ReLU neuron models are used in this part. In the case of convolutional and recurrent network, hidden layers with Long Short Term Memory (LSTM) neuron models, or Gated Recurrent Units (GRU) are used on top of the CNN part, providing the final classification, or prediction, outputs. 

In our approach we select to extract and further analyse the, say $M$, outputs of the last fully connected  layer, or last hidden layer of the trained CNN, or CNN-RNN respectively.  This is due to the fact that these outputs constitute high level, semantic extracts, based on which the trained DNN provides its final predictions. Other choices involve features extracted, not only from high level, but also from mid and lower level layers. In the following we present the extraction of concise semantic information from these representations, using unsupervised analysis.

Let us assume that a dataset \textit{S}, including medical input images has been collected and used for training a DNN to  predict the healthy, or not healthy, status of subjects. Let also  $T$  denote the respective test set used to evaluate the performance of the trained network. We train the DNN using the data in $S$, with cardinality $N_{s}$ and, for each input $k$, we collect the $M$ values of the outputs of neurons in the selected DNN fully connected or hidden layer, generating a vector ${\textbf{v}}_{s}(k)$. A similar vector  ${\textbf{v}}_{t}(k)$ is generated when applying the trained DNN to each input $k$ of the $N_{t}$ test set :

\begin{equation}
\label{eq:traininglatent}
\mathcal{V}_s = \big\{({\textbf{v}}_{s}(k), \ k=1,\ldots,N_{s}\big\} 
\end{equation}
and 
\begin{equation}
\label{eq:testlatent}
\mathcal{V}_t = \big\{({\textbf{v}}_{t}(k),  \ k=1,\ldots,N_{t}\big\} 
\end{equation}

In the following we derive a concise representation of these $\textbf v$ vectors, by using a clustering procedure, such as the k-means++ algorithm  \cite {ref20} to generate, 

\noindent
say, $L$ clusters  ${Q} =\{\textbf{q}_1,\ldots,\textbf{q}_L\}$ through minimization of the following function: 
\begin{equation}
\label{eq:kmeans}
\widehat{{Q}}_{k\text{-means}} = \underset{{Q}}{\operatorname{arg\ min}} 
\sum_{i=1}^{L} \sum_{\mathbf{v}_{s}\in {V}_{s}}^{} 
\big|\big|\textbf{v}_{s}-\textbf{$\mu$}_{i}\big|\big|^{2}
\end{equation}
in which $\textbf{$\mu$}_{i}$ denotes the mean of $\textbf v$ values belonging to cluster $i$. 
For each cluster $i$, we then compute the corresponding cluster center $\textbf{c}(i)$, thus defining the set of cluster centers $C$, which forms a concise representation and prediction model for medical diagnosis.

\begin{equation}
\label{eq: cluster centroid set}
\mathcal{C} = \big\{(\textbf{c}(i), \ i=1,\ldots,L\big\} 
\end{equation}

\begin{figure}[t]
    \begin{center}
        \includegraphics[scale=0.3]{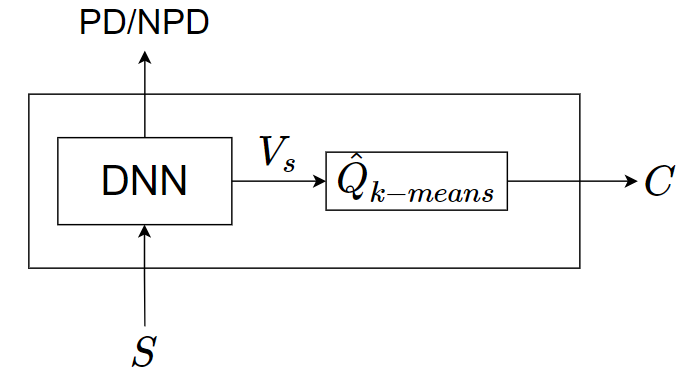}
        \caption{Dataset $S$ is the DNN input; clustering of $\mathcal{V}_s$ produces representation $C$}
              \setlength{\belowcaptionskip}{-2cm} 
        \label{fig:fig3}
    \end{center}
\end{figure}

The procedure, of using dataset $S$ to generate the set of cluster centers $C$ is illustrated in Fig.~\ref{fig:fig3}. Since the derived representation consists of a small number of cluster centers, medical experts can examine and annotate the respective DaTscans and MRI images with relevant textual information. This information can include the subject's status, as well as other metrics. Let us now focus on using the set $C$ for diagnosis in new subject cases, e.g., those included in the test dataset $T$. For each input in $T$, we compute the 
$\textbf{v}_{s}$ value. We then calculate the euclidean distance of this value from each cluster center in $C$
and classify it to the category of the closest cluster center. As a result, we classify each test input to a respective category, thus predicting the subject's status. 

It should be mentioned that, using this approach, we can predict a new subject's status in a rather efficient and transparent way. At first, only $L$ distances between $M$-dimensional vectors have to be computed and the minimum of them  be selected. Then, the subject can be informed of why the specific diagnosis was made, through visualization of the medical images and presentation of the medical annotations corresponding to the selected cluster center. 

\vspace{-0.2cm}
\subsection{The Unified Prediction Model}

Following the above described approach: a) we design a DNN (as shown in Fig.\ref{fig:fig3}) and extensively train it for predicting a disease, based on image data provided by a specific hospital, or available database, b) we generate a concise representation (set $C$) composed of the derived cluster center representation that can be used to predict the disease in an efficient way. This
information, i.e., the DNN weights and the set $C$, represent, in the proposed unified approach, the knowledge obtained through the analysis of the respective database $S$.

\begin{figure}[t]
    \begin{center}
        \includegraphics[scale=0.30]{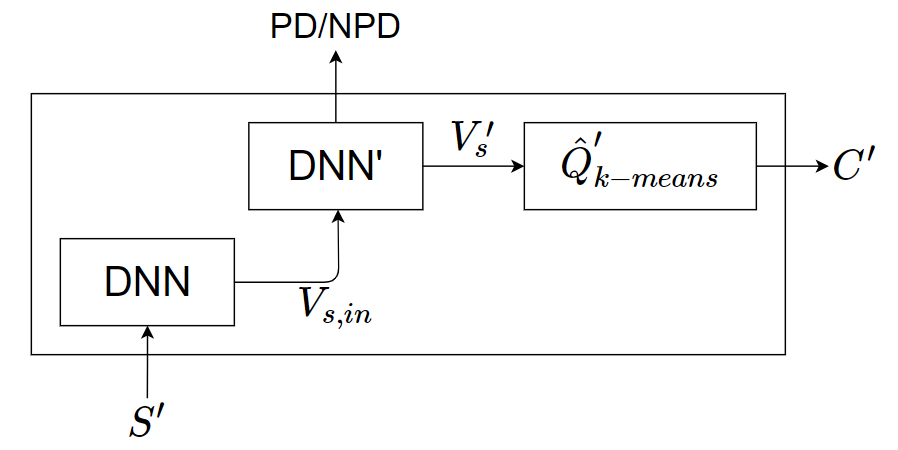}
               \setlength{\belowcaptionskip}{-0.7cm} 
        \caption{$S'$ is DNN input; ${V}_s$ is DNN' input; clustering ${V'}_s$ produces representation $C'$}
        \label{fig:fig4}
    \end{center}
\end{figure}

Let us now consider another medical environment, where another database related to the disease has been generated. Let us assume that it can be, similarly, described through the respective $S$ and $T$ sets. 
We will show how the proposed approach can alleviate the `catastrophic forgetting' problem. 
Fig. \ref{fig:fig4} shows the procedure we follow to achieve such a model. According to it, we present all inputs of the new training dataset $S'$ to the available DNN that we have already trained with the original dataset $S$; we compute the $\mathcal {V}_{s}$ representations, similarly to (3), named as $V_{s, in}$ in Fig. \ref{fig:fig4}. These representations, which were generated using the knowledge obtained from the original dataset, form the input to a new DNN, named DNN' in Fig. \ref{fig:fig4}; this network is trained  to use these inputs so as to predict the PD/NPD status of the subjects whose data are in set $S'$.  

\begin{equation}
\label{eq: cluster centroid set2}
\mathcal{C'} = \big\{(\textbf{c'}(i), \ i=1,\ldots,L'\big\} 
\end{equation}

\begin{figure}
    \begin{center}
        \includegraphics[scale=0.30]{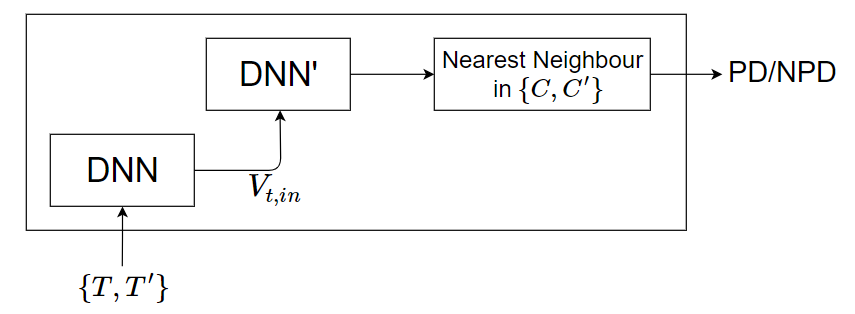}
              \setlength{\belowcaptionskip}{-0.7cm} 

        \caption{Sets $T$, $T'$ are DNN-DNN' inputs; $\mathcal{V'}_s$ classified through $C$ and $C'$}
        \label{fig:fig5}
    \end{center}
\end{figure}

In a similar way, as in Eqs. (3)-(5), we compute the new set of representations, named $V'_{s}$ and through clustering the new set of cluster centers $C'$:
The next step is to merge the sets $C$ and $C'$, creating the unified prediction model. Using the two network structures (DNN and DNN' in Fig. \ref{fig:fig4}), in a testing formulation, and the nearest neighbor criterion with respect to the union of $C$ and $C'$, we can predict the PD/NPD status of all subjects in both test sets {T} and {T'}, as  shown in Fig. \ref{fig:fig5}.  
The resulting representation, consisting of the $C$ and $C'$ sets, can predict a new subject's status, using the knowledge acquired by DNN and DNN' networks trained on both datasets, in an efficient and transparent way.   

In summary, we have trained the original DNN over a well defined input data space (i.e., large number of medical images, balanced data categories, etc). Then we do not train the new DNN’ with the respective (new) medical image data set, but with the concise representations - thus achieving lower dimensionality and lower risk of overfitting. Moreover, we take advantage of the knowledge of the originally trained DNN (which provides the input to DNN’). In this way, knowledge is scaled and interweaved between the two networks; it is not simply accumulated at their outputs.

\section{Domain Adaptation from multiple sources}

In this Section we describe the potential extension of the unified prediction framework generated with the annotated medical datasets, to other non-annotated datasets. We are currently implementing a pipeline which includes generation of the latent representation sets for COVID-19 diagnosis and uses DA to extend this framework to new datasets, which are not annotated. Since this pipeline is currently under implementation, we briefly sketch how we will implement DA from multiple sources in this framework, using optical character verification in retail food packaging~\cite{ref21} as an example.

In particular, we will adopt the Domain Adaptation approach developed in \cite{ref33} for using labeled source data from, say, $N$  environments, so as to provide annotation of unlabeled data in a target environment.  

The model comprises a feature extractor and a classification part. The feature extractor part learns useful representations from all environments/domains - such as the latent variables representations derived above, whereas its sub-network learns features specific to each source-target domain pairs. The classification part of the model learns domain-specific attributes for each target image and provides, say, N categorization results. It aligns with domain-specific classifiers, as the class boundaries are highly likely to be misclassified, because they are learned from different classifiers. 

The model aims at minimizing: (i) the feature discrepancy, for learning domain invariant representations, (ii) the class boundary discrepancy, for minimizing the mismatch among all classifiers, (iii) the classification loss, improving source data classification; consequently, leading to improved generalization on the target data set. The sum of these three Losses is the overall objective function of architecture training. A Class Activation Mapping (CAM) module is included to visualize the DNN’s interpretation when making predictions. As is well-known, CAM highlights which areas of the input images contribute the most to the DNN decision making process.

In this approach, feature discrepancy is reduced by minimizing both MMD and CORAL loss. 
MMD defines the distance between the two data distributions. 
CORAL loss is also used to minimize the discrepancy
between source and target data by reducing the (covariance) distance
between the source and target feature representations. 

Classifiers are likely to misclassify
the target samples near the class boundary as they are
trained using different source domains, each having different
target prediction. Therefore it is aimed to minimize discrepancy
among, say, $N$ classifiers by making their probabilistic outputs
similar. 
The network also reduces the discrepancy
among classifiers by minimizing the classification loss. The network is trained 
with labeled source data and minimizes the cross-entropy loss.

The total loss is made up of classification loss, feature
discrepancy loss and class discrepancy loss. By
minimizing this, the network can classify
source domain data more accurately and reduce dataset
bias and discrepancy among classifiers. 

\vspace{-0.2cm}
\section{Experimental Study}

 In the following, we apply the proposed prediction approach, based on latent information extraction, to two different Parkinson’s datasets (one created in collaboration with the Georgios Gennimatas Hospital in Athens and the PPMI study sponsored by M. J. Fox for Parkinson’s Research) including MRI and DaTScans for Parkinson’s prediction. Next, we apply the proposed approach for automatically detecting COVID-19 using chest CT scan series and chest x-rays. The most challenging part of this task is to detect the small nodules and lesions in the early stage of the COVID-19 and differentiate it from normal cases or pneumonia. The datasets of RSNA pneumonia detection and LUNA detection are used as source datasets, enriched by smaller datasets that exist for COVID-19. Finally, the proposed approach including domain adaptation is applied to the optical verification of end-by date on food retail packaging. In all cases the proposed approach improves the current state-of-the-art, providing a transparent and easy to implement procedure for prediction of diseases, or anomalies, based on analysis of real images aggregated in the related environments.    

\vspace{-0.2cm}
\subsection{Prediction of Parkinson's based on MRI and DaTscans}

In \cite{ref2}, DNNs were trained with an augmented dataset based on the Greek database, achieving very good performance on this database. The  convolutional part of the network was applied to each image component, i.e., to an RGB DaTscan image and to three (gray-scale) MRIs,  using the same pretrained ResNet-50 structure. The outputs of these two ResNet structures were concatenated and fed to the Fully Connected (FC) layer of the CNN part of the network. 

This structure has been able to analyse the spatial characteristics of the DaTscans and MRIs, achieving a high accuracy in the database test set, of 94\%, as shown in Table~\ref{table:table1}. 
The complete CNN-RNN architecture included two hidden layers on top of the CNN part, each containing 128 GRU neurons, as shown in the Table. This has been able to also analyse the temporal evolution of the MRI data, achieving an improved performance of 98 \% over the test data. 
We trained this CNN-RNN network so as to classify the DaTscans and MRIs to the correct PD/NPD category,  using a batch size of 10, a fixed learning rate of 0.001 and a dropout probability of 50 \%.  The clustering process, using k-means was then applied to the $\mathcal{V}_{s}$ vectors. Based on best precision/recall over the generated clusters, we extracted five clusters, two of which correspond to control subjects, i.e. NPD ones, with three clusters corresponding to patients. These constitute the extracted concise representation $C$ set; consequently, $C$ is composed of five 128-dimensional vectors.  

\begin{table*}[!t]
\caption{The accuracy obtained by CNN and CNN-RNN architectures}
\label{table:table1}
\centering
\begin{tabular}{|c|c|c|c|c|c|}
\hline 
Structure & FC & Hidden & Units in FC & Units in Hidden & Accuracy ($\%$) \\
& Layers & Layers & Layers & Layers &\\
\hline \hline
CNN & 2 & - & 2622-1500 & - & 94 \\
CNN-RNN & 1 & 2 & 1500 & 128-128 & 98 \\
\hline 
\end{tabular}
\end{table*}

\begin{figure}
    \begin{center}
        \includegraphics[scale=0.25]{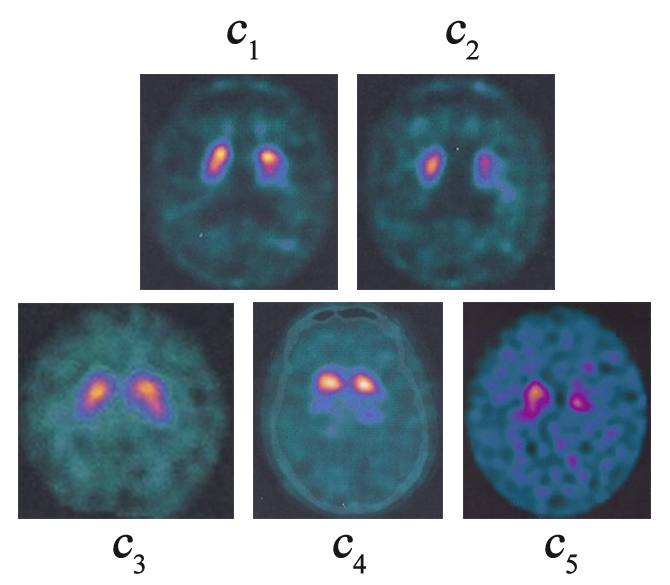}
       \caption{The DaTscans of the 5 selected cluster centers: $\textbf{c}_{1}$ and $\textbf{c}_{2}$ correspond to NPD cases, whilst $\textbf{c}_{3}$ - $\textbf{c}_{5}$ to progressing stages of Parkinson's}
        \label{fig:fig8}
    \end{center}
\end{figure}

The DaTscans corresponding to the extracted cluster centers are shown in Fig.~\ref{fig:fig8}. Through the assistance of  medical experts we were able to verify that the three DaTscans corresponding to patient cases represent different stages of Parkinson's disease. In particular: the first of them ($\textbf{c}_{3}$) represents an early occurrence, between stage 1 and stage 2; the second ($\textbf{c}_{4}$) shows a pathological case, at stage 2; the third ($\textbf{c}_{5}$) represents a case that has reached stage 3 of Parkinson's. In the case of controls, there are differences between the first ($\textbf{c}_{1}$), which is a clear NPD case and the second ($\textbf{c}_{2}$), which is a more obscure case.

Following the above annotations, it can be said that the derived representations convey more information about the subjects' status than trained DNN outputs. This information can be used by medical experts to evaluate the predictions made by the original DNN when new subjects' data have to be analysed. The computed $\mathcal{V}_{s}$ representations in the new cases can be efficiently classified to the category of the nearest cluster center of $C$; the cluster center's Datscan, MRIs and annotations will then be used to justify, in a transparent way, the provided prediction. In case of new data, retraining of the deep neural network would  be required, so as to retain the old knowledge and include the new one; this would be computationally intensive and possibly unfeasible. On the contrary, the proposed approach would only require extension of the $C$ set with one, or more, cluster centers, corresponding to the new information; as a consequence, this would be done in a very efficient way.

Next, we examine the ability of the procedure shown in Fig. \ref{fig:fig4}, using the trained DNN (CNN-RNN) architecture,  to be successfully applied to the PPMI database \cite{ref11}, for PD/NPD prediction. Since the DaTscans were the basic source of the DNN's discriminating ability, we focus our new developments on the DaTscans included in the PPMI database. For this reason, we have retained 609 subjects from the PPMI database, excluding some patients for which we were not able to extract DaTscans of good quality. In total we selected 1481 DaTscans, which we combined with MRI triplets from the respective subjects, generating a dataset of 7700 inputs; each input was composed, of one (gray-scale) DaTscan and a triplet of MRI images.

At first, for comparison purposes, we trained CNN and CNN-RNN networks, similar to the ones presented in the previous subsection, from scratch, on the selected PPMI training set (6656 inputs). We used the validation set (1584 inputs) to test the obtained accuracy in the end of each training epoch. We then  tested the performance of the networks on the test set (2028 inputs). The obtained accuracy was in the range of 96-97 \%, similar to the accuracy achieved by the state-of-the-art techniques on PPMI. We also used TL on networks generated in the first subsection of our experimental study, to initialise the re-training of the new networks. Similar results were obtained in this case as well.

We then applied the procedure shown in Fig. \ref{fig:fig4}, to train DNN' with the $\mathcal{V}_{s}$ vectors extracted from the last hidden layer of the DNN that had been trained on \cite{ref10,ref34}. 
The performance of the network was very high, classifying in the correct PD/NPD category 99.76 \% of the inputs. 
\begin{figure}
    \begin{center}
        \includegraphics[scale=0.25]{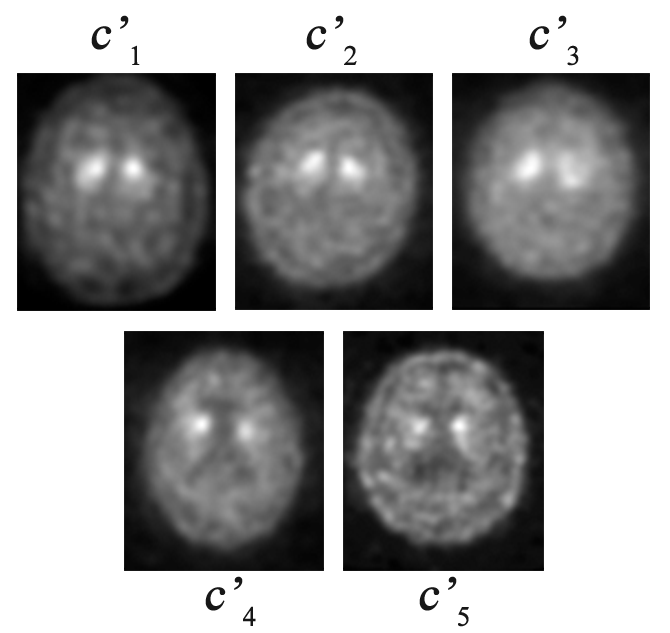}
    \caption{DaTscans of the 5 cluster centers in $C'$, with NPD at top and PD at bottom}
        \label{fig:fig12}
    \end{center}
\end{figure}
By then implementing the clustering procedure shown in Fig. \ref{fig:fig4}, we were able to extract five new clusters, three of which represent NPD subjects' cases and two of which represent PD cases. These cluster centers are 32-dimensional vectors. Fig. \ref{fig:fig12} shows the DaTscans corresponding to the cluster centers $\textbf{c'}_1$ - $\textbf{c'}_5$. Since the patients in the PPMI Database generally belong to early stages of Parkinson's (stage 1 to stage 2), it can be seen that two cluster centers, i.e., $\textbf{c'}_4$ and $\textbf{c'}_5$ were enough to represent these cases. Variations in the appearance of the non-Parkinson's cases can be seen in $\textbf{c'}_1$ - $\textbf{c'}_3$ DaTscans.

We then applied the merging of sets $C$ and $C'$. It should be mentioned that the 5 centers in set $C$ were 128-dimensional, whilst the 5 centers on set $C'$ were 32-dimensional. To produce a unified representation, we made an ablation study, through PCA analysis, on the classification performance achieved in dataset \cite{ref10}, if we represented the five cluster centers in $C$ through only 32 principal components. We were able to achieve a classification performance of 97.92 \%, which is very close to the 98 \% performance in Table \ref{table:table1}.  
\begin{figure}[t]
    \begin{center}
        \includegraphics[scale=0.4]{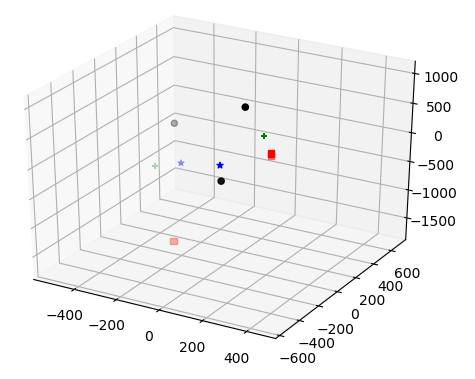}
       \caption{The obtained ten cluster centers in 3-D: 5 of them (squares with red/rose color, \&  plus (+) symbols with green color) depict patients; 5 of them (stars with blue color \& circles with black/grey color) depict non-patients}
        \label{fig:fig13}
    \end{center}
\end{figure}
Consequently, we were able to generate a unified model consisting of 10 32-dimensional cluster centers. Fig. \ref{fig:fig13} shows a 3-D projection of the ten cluster centers. The three (red/rose) squares denote the patient cases in the dataset \cite{ref10} and the two (green) plus (+) symbols represent the patient cases in the PPMI dataset. The two (blue) stars represent the normal cases in the Greek dataset and the three (black/grey)  circles represent the normal cases in the PPMI dataset. It can be seen that the PD centers are distinguishable from the NPD ones.

This has been  verified by testing the ability of the unified prediction model to correctly classify all input data in test sets $T$ and $T'$, i.e., the data from both datasets. There was no effect on the performance of the prediction achieved by each prediction model, i.e., $C$ and $C'$ when applied, separately, to their respective datasets. 
This shows that the unified representation set, composed of the union of $C$ and $C'$, has been able to provide exactly the same prediction results, as the original representation sets. 

Let us further discuss the significance of the derived cluster centers, for generating trustworthy DNN decision making in healthcare. Whenever a PD/NPD prediction is provided to the medical expert for a specific subject, it will also show the subject's DaTscan, together with the DaTscan of the center of the selected cluster. The latter will  indicate what type of data were used by the system to generate its prediction. In this way, the medical expert, and the subject, could decide by themselves whether to trust, or not, the suggested decision.

\vspace{-0.3cm}
\subsection{Prediction of COVID-19 based on CT chest scans and x-rays}

We applied the proposed procedure for detection of COVID-19, based on two medical image types, i.e., CT chest scans and chest x-rays.

The COVID19-CT dataset \cite{ref25}, is an open source public dataset, containing 349 CT scans of 143 patients positive for COVID-19 and 397 CT scans of people that are negative for COVID-19 (subjects that are normal, or have other types of diseases). The images of positive patients are collected from medRxiv and bioRxiv papers about COVID-19. The authors selected CTs containing COVID-19 abnormalities by reading the figure captions in the papers and also manually removed artifacts in the original images. The CTs of people negative for COVID-19 are selected from the PubMed Central search engine and from the MedPix, a publicly-open online medical image database that contains CT scans with various diseases. The COVID19-CT dataset is split into training, validation and test sets which contain 191, 60, 98 CT scans positive for COVID-19 and 234, 58, 105 CT scans negative for COVID-19, respectively.

\begin{figure}
    \begin{center}
        \includegraphics[height = 2.2cm]{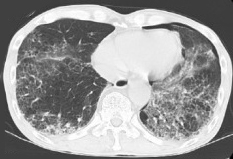}
        \includegraphics[height = 2.2cm]{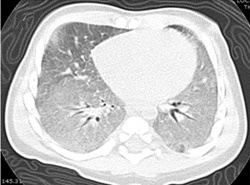}
        \includegraphics[height = 2.2cm]{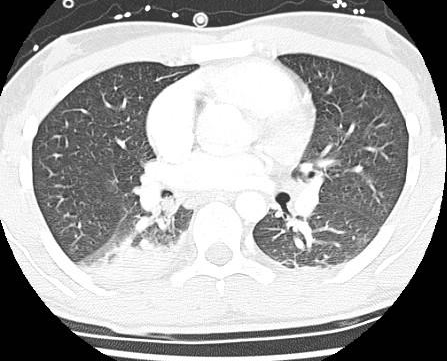}
        
        \includegraphics[height = 2.2cm]{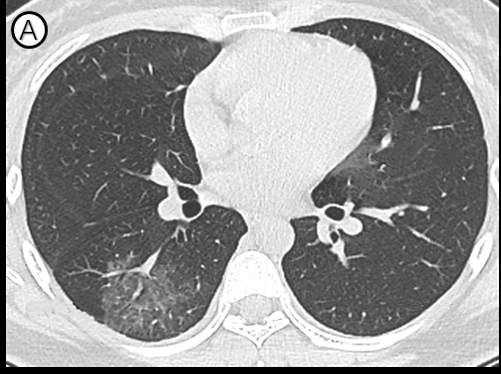}
        \includegraphics[height = 2.2cm]{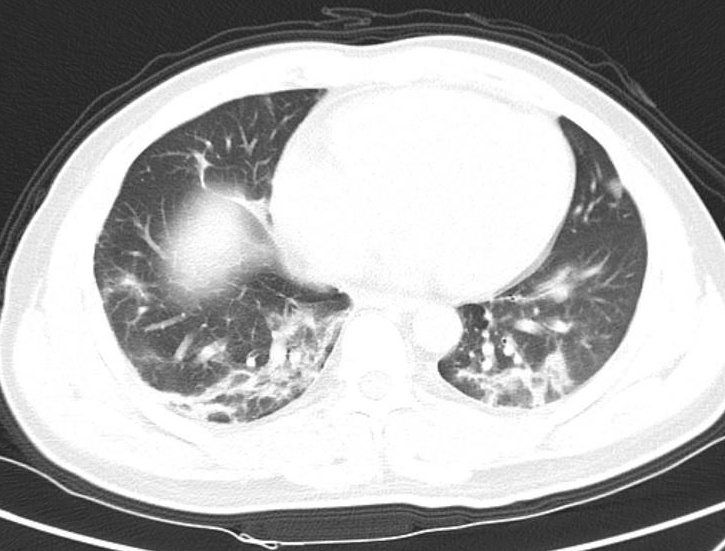}
    \setlength{\belowcaptionskip}{-1cm} 
    \caption{CT Scans of the 5 cluster centers: non-COVID at top,  COVID at bottom}
   
        \label{fig:ctscans}
    \end{center}
\end{figure}

In \cite{ref25} the EfficientNet-b0, pretrained on Imagenet (including 5,288,548 parameters), was retrained and tested on this dataset, providing an F1 score of 0.78. Data augmentation, based on random cropping,  horizontal flip, color jittering with random contrast and random brightness was used.
We applied the proposed procedure for latent variable extraction, using the same EfficientNet-b0 network, adding an extra hidden layer of 32 units, so as to extract 32-dimensional representations and using vertical and horizontal flipping for data augmentation. 5 clusters were generated, with their respective centroids shown in Fig. \ref{fig:ctscans}. The obtained F1 score was much higher 0.842 (0.855 for the non-COVID case and 0.828 for the COVID one), whilst providing the CT scans that are the (cluster center) attractors for new predictions.  
It should be mentioned that in \cite{ref25} another much more complex network, DenseNet-169 (with 14,149,480 parameters;  trained with additional images from the Lung Nodule Analysis dataset) achieved an F1 score of 0.85. 

In the case of chest x-rays, we used two different datasets, i.e., the COVID ChestXray dataset \cite{ref26} and the Kaggle RSNA pneumonia dataset. The first dataset consists of data compiled from websites such as Radiopaedia.org, the Italian Society of Medical and Interventional Radiology and Figure1.com. The second dataset is part of the RSNA Pneumonia Detection Challenge dataset. In total, the merged dataset consisted of images of subjects with pneumonia, subjects with COVID-19 and normal ones. The merged dataset was split into training and test parts, which contained 8629 and 955 pneumonia instances, 128 and 14 COVID-19 instances and 9766 and 885 normal instances, respectively.

\begin{figure}
    \begin{center}
        \includegraphics[height = 2.5cm]{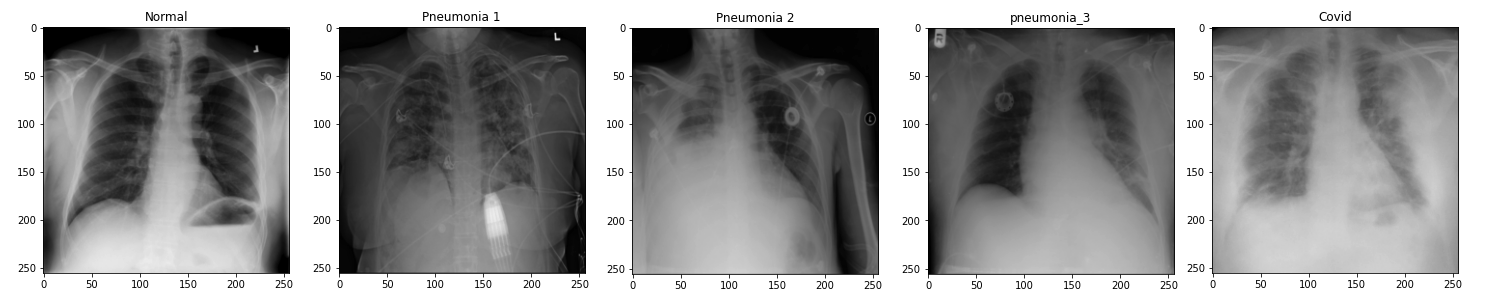}
    \caption{Chest x-rays of the 5 cluster centers}
        \label{fig:xrays}
    \end{center}
\end{figure}

In a work\footnote{https://github.com/velebit-ai/COVID-Next-Pytorch}, the COVID-Next network, which is a ResNeXt50-32x4d, was trained on this dataset. Data augmentation, based on  vertical and horizontal flip, affine transformations (translation, scaling, shearing) and color jittering was used. COVID-Next achieved an F1 score of 0.93.
We applied the proposed procedure, using the same network and transforms, adding an extra layer with 32 units. Five clusters of the derived representations were generated, with the respective centroids being shown in Fig. \ref{fig:xrays}. The achieved F1 score was 0.96, much higher than the state-of-the-art with the same network. 

We have further unified the extracted representations from CT scans and x-rays, verifying the improved classification achieved through the proposed approach.

\vspace{-0.6cm}
\subsection{Optical Character Verification in Retail Food Packaging}

The Food Packaging Image dataset used in this study consisted
of more than 30,000 balanced images (OK vs NOT-OK)
from six different locations in UK, and the target was to automatically verify the quality of
use-by dates printed on them. Initially, three people manually annotated
this dataset, with two more annotators further sampling and
verifying the manual annotations for quality control, hence
keeping those 30,000 images that both of them were in full agreement. The main challenges of these datasets were
unavailability of labeled data and high variability across
the datasets, such as heavy distortion, varying background,
illumination/blur, date format, angle and orientation of the
label. Training was carried out on a 70 \% of the samples
with another 10 \% used for validation. Finally, the
remaining 20 \% of the images were used for evaluating and
testing  across the six locations, in order to automatically verify the quality of printed
use-by dates, hence detecting images of very low quality.

Experiments
were conducted \cite{ref33}, using  the multi-source
DA approach, i.e., using two labeled
source datasets for adaptation to a single unlabeled target domain.
The obtained results were compared to the baseline single
source adaptation experiment conducted initially.  
The labeled sources and the unlabeled target images have
been fed through the model where the discrepancy between
the pair of datasets was minimized by jointly reducing 
feature discrepancy, class discrepancy and classification
losses. 
The multi-source
DA approach significantly outperformed single-source DA, 
with an average classification accuracy improvement by
more than 6 \%. In particular, the achieved performance through multi-source DA was very high, reaching an accuracy of 90.53 \%, compared to an accuracy of 84.14 \% in the single source DA case.  

As was above-mentioned, our current work includes application of the multi-source DA approach to COVID-19 prediction.

\section{Conclusions and Future Work}

In this paper we have developed a new approach for deriving efficient and transparent prediction models and used them for prediction of Parkinson's, of COVID-19 and for detection of anomalies in optical character verification.  A crucial issue in the above described approach is related to the uncertainty introduced in DA, since labels are not  known when dealing with the new data. We are currently extending the developed DNN architectures using Bayesian formulations to provide self-training capabilities. We are using a sample-wise weighting scheme during training that places a weight on each training sample, according to the estimated uncertainty over predicted label, so that we incrementally encourage the model to assign more weight to uncertain label samples as training progresses \cite{ref9}.  

We are also working on representing the DA procedure as a multi-objective optimization problem \cite{ref15}. We are investigating the use of capsules in the developed algorithms \cite{ref13}.
In the future we will combine the data driven DNN representations with knowledge-based ontological representation and visual attention mechanisms  \cite{ref22,ref24,ref35,ref14,ref23} so as to formally explain the derived predictions.

GRNET is the main consultant of the Greek Ministry of Digital Governance (DG). Through their Harmoni project with the Ministry of Health, all radiological exams in public hospitals are directly transferred and stored in GRNET data centers. The Harmoni database does not include annotated medical exams. We will infer annotations and generate an enriched version of Harmoni. This will be an issue to discuss through the collaboration with CLAIRE Network that is supported by the Greek Ministry of DG and GRNET and with TAILOR Network to which NTUA and GRNET participate.  

%
%
%
\bibliographystyle{splncs04}
\bibliography{mybibliography}

\end{document}